\documentclass[11pt,a4paper]{article}
\usepackage[hyperref]{acl2020}
\usepackage{times}
\usepackage{latexsym}

\usepackage{multirow}
\usepackage{float}

\usepackage{microtype}

\aclfinalcopy 
\def\aclpaperid{17} 

\usepackage{subcaption}
\usepackage{booktabs}

\usepackage{graphicx}
\usepackage{multirow}
\usepackage{float}
\usepackage{placeins}
\usepackage{tikz}

\title{What Gives the Answer Away? \\
Question Answering Bias Analysis on Video QA Datasets}

\author{
    Jianing Yang$^1$\quad 
    Yuying Zhu$^2$\quad
    Yongxin Wang$^3$\quad
    Ruitao Yi$^4$\quad\\
    \bf Amir Zadeh$^2$\quad
    Louis-Philippe Morency$^2$\\
    $^1$Machine Learning Department, Carnegie Mellon University \\
    $^2$Language Technologies Institute, Carnegie Mellon University \\
    $^3$Robotics Institute, Carnegie Mellon University \\
    $^4$Department of Electrical and Computer Engineering, Carnegie Mellon University \\
    \texttt{\{jianing3,yuyingz,yongxinw,ruitaoy,abagherz,morency\}@cs.cmu.edu}\\}

\date{}

\begin{document}

\maketitle

\begin{abstract}
Question answering biases in video QA datasets can mislead multimodal model to overfit to QA artifacts and jeopardize the model's ability to generalize. Understanding how strong these QA biases are and where they come from helps the community measure progress more accurately and provide researchers insights to debug their models. In this paper, we analyze QA biases in popular video question answering datasets and discover pretrained language models can answer 37-48\% questions correctly without using any multimodal context information, far exceeding the 20\% random guess baseline for 5-choose-1 multiple-choice questions. Our ablation study shows biases can come from annotators and type of questions. Specifically, annotators that have been seen during training are better predicted by the model and reasoning, abstract questions incur more biases than factual, direct questions. We also show empirically that using annotator-non-overlapping train-test splits can reduce QA biases for video QA datasets.
\end{abstract}

\section{Introduction}

Video understanding is a central task of artificial intelligence that requires complex grounding and reasoning over multiple modalities. Among many tasks, multiple-choice question answering has been seen as a top-level task \citep{richardson_mctest_2013} toward this goal due to its flexibility and ease of evaluation. A line of research towards constructing Video QA datasets have been completed \cite{Tapaswi2016MovieQAUS,Lei2018TVQALC,Zadeh2019SocialIQAQ}. Ideally, a model for this task should understand each modality well and have a good way to aggregate information from different modalities. To this end, it is a natural choice for researchers to use the state-of-the-art models for each subtask and modality. Recently in the Natural Language domain, BERT \citep{devlin-etal-2019-bert} and other transformer-based models have become baselines in many research works. However, it is a known phenomenon that complex multimodal models tend to overfit to strong-performing single modality \cite{cirik_visual_2018,mudrakarta_did_2018, thomason_shifting_2019}. To caution against such undesirable modality collapsing, we study how strong RoBERTa \citep{Liu2019RoBERTaAR}, a better trained version of BERT, can perform on the QA-only task.

\paragraph{Our main contribution includes:}
(1) Show that RoBERTa baselines exceed all previously published QA-only baselines on two popular video QA datasets. (2) The strong QA-only results indicate the existence of non-trivial biases in the datasets that may not be obvious to human eyes but can be exploited by modern language models like RoBERTa. We provide analyses and ablations to root-cause 
these QA biases, recommend best practices for dataset splits and share insights on subjectivity vs. objectivity for question answering.

\section{Model}
\label{sec:model}

\begin{table*}[h]
\centering
\begin{small}
\begin{tabular}{@{\hspace{1pt}}c@{\hspace{20pt}}l@{\hspace{20pt}}l@{\hspace{20pt}}l@{\hspace{5pt}}c}
    \bf Dataset & \bf Model Name/Source & \bf Modality & \bf QA Model & \bf Val Acc (\%) \\
    \toprule
    \multirow{4}{*}{\shortstack{MovieQA \\(A5)}}   & \bf Our Answer-only & \bf A      & \bf RoBERTa (fine-tune)    & \bf 34.16\\
                                                   & \bf Our QA-only & \bf Q+A      & \bf RoBERTa (fine-tune)    & \bf 37.33\\
                                                   & \bf Our QA-only& \bf Q+A      & \bf RoBERTa (freeze) & \bf 22.52\\
                                                   &  SOTA \cite{Jasani2019AreWA}   &V+S+Q+A &  w2v         &  48.87\\
                                                   &  Random Guess    &     -   & -     & 20.00 \\
                                    
    \cmidrule{1-5}
    \multirow{6}{*}{\shortstack{TVQA \\(A5)}}      & \bf Our Answer-only & \bf A      & \bf RoBERTa (fine-tune)    & \bf 46.58\\
                                                   & \bf Our QA-only& \bf Q+A      & \bf RoBERTa (fine-tune) & \bf 48.91\\
                                                   & \bf Our QA-only& \bf Q+A      & \bf RoBERTa (freeze) & \bf 30.75\\
                                                   & QA-only with Glove \cite{Jasani2019AreWA} & Q+A &  GloVe + LSTM & 42.77\\
                                                   & SOTA's QA-only \cite{yang2020bert}& Q+A &  BERT (fine-tune)         &  46.88 \\
                                                   & SOTA\cite{yang2020bert}   &V+S+Q+A&  BERT (fine-tune)        &  72.41    \\
                                                   &  Random Guess      &  -     & -    &  20.00    \\

	\bottomrule
\end{tabular}
\end{small}
\caption{
Comparison with State-of-the-art Performance.
}
\label{tab:baselines}
\end{table*}

\begin{table*}[]
\centering
\begin{small}
\begin{tabular}{|l|c|c|c|c|c|c|}
\hline
                   & \multicolumn{1}{l|}{\textbf{\# of questions}} & \multicolumn{1}{l|}{\textbf{\# of annoators}} & \multicolumn{1}{l|}{\textbf{\% of why/how }} & \multicolumn{1}{l|}{\textbf{\% of other type }} & \multicolumn{1}{l|}{\textbf{avg len of Q}} & \multicolumn{1}{l|}{\textbf{avg len of A}} \\ \hline
\textbf{Movie QA}  & 14,944                                       & ---                                           & 20.9\%                                                & 79.1\%                                                   & 5.2                                                      & 5.29                                                \\ \hline
\textbf{TVQA}      & 152,545                                      & 1,413                                          & 14.5\%                                                & 85.5\%                                                   & 13.5                                                  & 4.72                                                \\ \hline
\end{tabular}
\end{small}
\caption{Dataset Statistics.}
\end{table*}

We fine-tune pretrained RoBERTa from \citet{Liu2019RoBERTaAR} to solve the question answering task. Specifically, for one multiple-choice question with five answers (1 correct and 4 incorrect), we concatenate the tokenized question with each of the five tokenized answers and feed each of these five q-a pairs into RoBERTa. The RoBERTa is connected with a 4-layer MLP (Multi-Layer Perceptron) head to produce a scalar score for each q-a pair. These five scores are then passed through Softmax to output five probabilities indicating how likely the model think it is for each q-a pair to be correct. During training, the probabilities are trained on Cross Entropy loss; during testing, the q-a pair with the highest probability is selected as the model's prediction.


\section{Datasets}
\label{sec:datasets}

We evaluate our baseline model against two popular multimodal QA datasets: MovieQAand TVQA.\\

\paragraph{MovieQA:}
MovieQA \citep{Tapaswi2016MovieQAUS} was created from 408 subtitled movies. Each movie has a set of questions with 5 multiple choice answers, only one of which is correct. The dataset also contains plot synopses collected from Wikipedia.

\paragraph{TVQA:} 
TVQA \citep{Lei2018TVQALC} was collected from 6 long-running TV shows from 3 genres. There are 21,793 video clips in total for QA collection, accompanied with subtitles and aligned with transcripts to add character names. Depending on the type of TV shows, a video clip is in 60 or 90 seconds. Each video clip has a set of questions with 5 multiple choice answers, only one of which is correct.

\paragraph{Notation:}
In this paper, we use A5 to denote the tasks on datasets. A5 means the multiple choice question consists of 1 correct answer and 4 incorrect answers.

\section{Bias Analysis}

\subsection{QA Bias and Inability to Generalize}

\begin{table}[h!]
\centering
\begin{small}
\begin{tabular}{@{\hspace{5pt}}l@{\hspace{10pt}}r@{\hspace{10pt}}r@{\hspace{10pt}}}
    \bf Train Set   & \multicolumn{2}{c}{{\bf Validation Accuracy (\%)}} \\
    \toprule
                    & \bf MovieQA   & \bf TVQA  \\
    \cmidrule{2-3}
    MovieQA         & \bf 37.33     & 31.18   \\
	TVQA            & 33.45         & \bf 48.91  \\

	\bottomrule
\end{tabular}
\end{small}
\caption{
Across-dataset generalization accuracy. Both datasets are trained and evaluated on the A5 task: multiple-choice questions with 1 correct answer and 4 incorrect answers (random guess yields 20\% accuracy). \textbf{Bold} number is the highest number in each column.
}
\label{tab:interdataset_generalization}
\end{table}

For the two datasets introduced in Section \ref{sec:datasets}, we perform QA-only baselines using pretrained language model as described in Section \ref{sec:model}. Table \ref{tab:baselines} shows how our QA-only model's performance compares to random guess, state-of-the-art full modality performance and its associated QA-only ablation performance.

From Table \ref{tab:baselines}, looking at the numbers in bold font,  we discover language model like RoBERTa is able to answer a significant portion of the questions correctly, despite that these questions are supposed to be not answerable without looking at the video. This result indicates that the model exploits the biases in these datsets. In addition, we also find that answer-only performance is quite close to QA-only performance, indicating the answer alone gives the model a pretty good hint on whether it is likely to be a correct answer.

Knowing there are biases in the datasets, we are then curious on if these learned biases are transferable between datasets. Tihs investigation is important because if the biases are transferable, then perhaps they are not necessarily bad, because one could argue the model has captured some common sense in these questions and answers; but if these biases are not transferable, then it means these biases only patterns tied to one particular dataset, which we hope the model not to learn. To verify this with experiments, we train a model on each of the two dataset's train split and evaluate these two models on each of the two dataset's validation split. The results are shown in Table \ref{tab:interdataset_generalization}.

Looking at each row in Table \ref{tab:interdataset_generalization}, we see all transfer-dataset evaluation's performance decreases from same-dataset evaluation. This means that although the model learns some tricks to answer the questions without context, such tricks learned from one dataset no longer works when applied at a different dataset. In other words, the model learns bias in the dataset and such bias is not transferable. This undesirable behavior is what motivates to our analysis in the next sections.

\subsection{Source of Bias: Annotator}

\newcommand{\decreased}[1] {\textcolor{red}{#1$\downarrow$}}
\newcommand{\increased}[1] {\textcolor{green}{#1$\uparrow$}}

\begin{table*}[h!]
\centering
\begin{small}
\begin{tabular}{c c c@{\hspace{4pt}} c@{\hspace{4pt}} c@{\hspace{4pt}} c@{\hspace{4pt}} c@{\hspace{4pt}} c@{\hspace{4pt}} c@{\hspace{4pt}} c@{\hspace{4pt}} c@{\hspace{4pt}} c@{\hspace{4pt}}}
    & \bf Overlap Acc (\%)  & \multicolumn{10}{c}{\textbf{Non-overlap Acc Shift (\%) vs. Dropped annotator}} \\
    \toprule
    \multirow{3}{*}{\shortstack{TVQA\\ (A5)}} & & w17 & w366 & w24 & w297 & w118 & w313 & w14 & w19 & w2 & w254 \\
    \cmidrule{3-12}
    & \bf 50.59 & \decreased{-5.59} & \decreased{-11.28} & \decreased{-20.14} & \decreased{-10.55} & \increased{+23.22} & \decreased{-20.59} & \decreased{-1.69} & \decreased{-5.96} & \decreased{-12.23} & \decreased{-17.28}\\

	\bottomrule
\end{tabular}
\end{small}
\caption{
Non-overlapping dataset re-split results on the top-10-annotator subset. The ``Overlap Acc" column is the validation accuracy where the train and validation set both contain questions from all 10 annotators. The ``Non-overlap Acc Shift vs. Dropped annotator" is the validation accuracy where the train set contains questions from 9 annotators and the validation set only contains questions from the dropped annotator.
}
\label{tab:non-overlap_resplit}
\end{table*}

\begin{figure}[t]

  \centering
  \includegraphics[width=\linewidth]{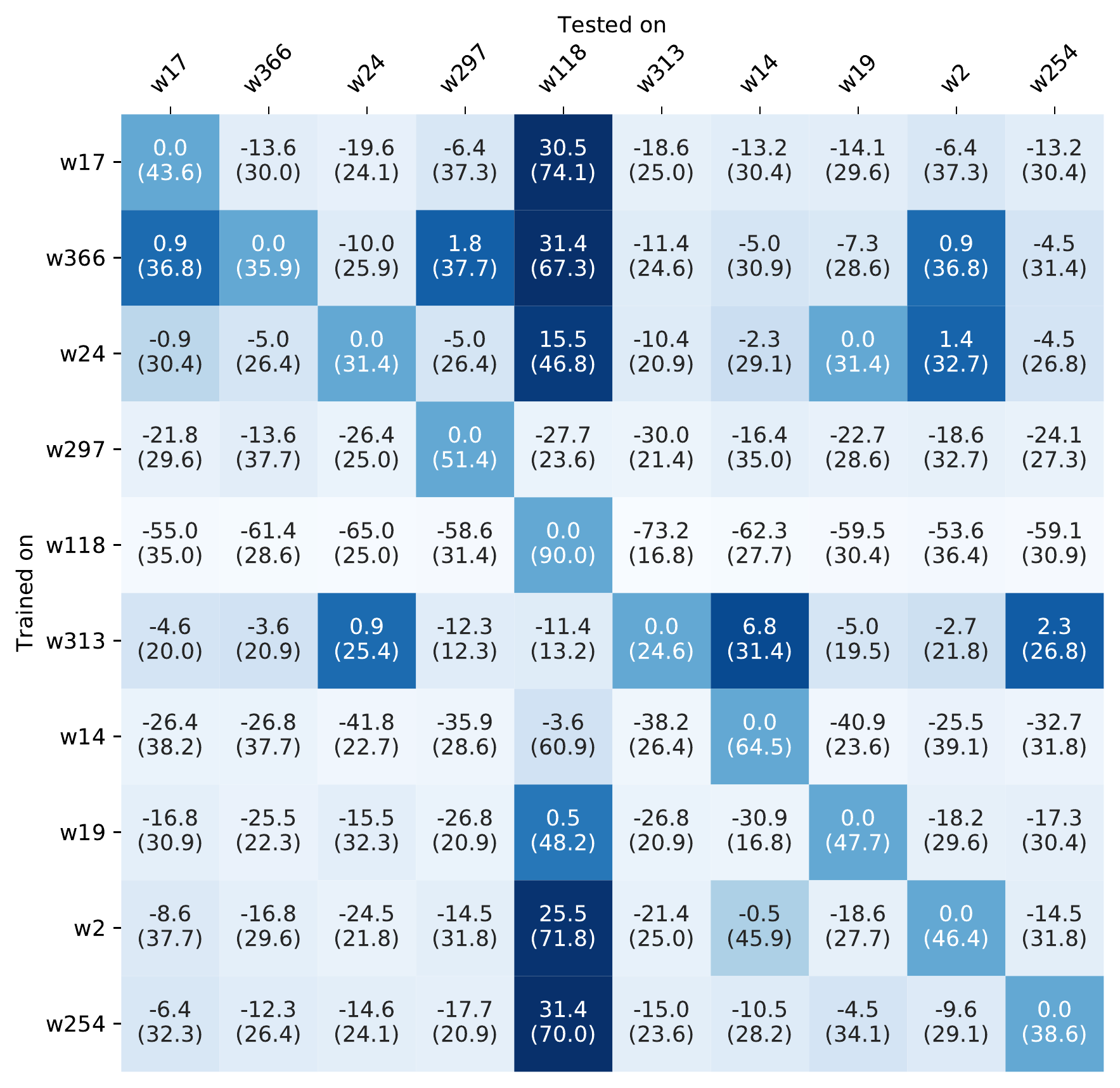}  
 
\caption{TVQA Inter-annotator accuracy shift confusion matrix. Each $w_i$ represents an annotator id and each cell represents a train-test combination between annotators. The cells are colored based on accuracy shift (the top number in each cell): lighter color means more negative accuracy shift and darker color means more positive accuracy shift. Accuracy shift is defined as the difference between each cell's accuracy (the bottom number) and the same-row diagonal cell's accuracy (again, the bottom number).}
\label{fig:inter_annotator}
\end{figure}

We hypothesize one source of bias is from annotators. To verify our hypothesis, we obtain the Annotator IDs corresponding to the questions in TVQA \footnote{We thank the authors of TVQA for sharing this information. Annotator information for MovieQA is unfortunately not available to us.} and construct a confusion matrix between the top-10 annotators. The results are shown in Figure \ref{fig:inter_annotator}. For each of the annotators, we construct a mini-train and mini-valid set. For TVQA, each mini-train and mini-valid set contains 1980 and 220 A5 questions, respectively.

Figure \ref{fig:inter_annotator} reveals a pattern where most cells except for those on the diagonals are light colored, which means the accuracy decreases when the train set's and validation set's questions are not from the same annotator. This indicates the model learns to guess for one specific annotator's questions but such guess strategy is not transferable to other annotator's questions. This reveals that RoBERTa has the capacity to overfit to the annotators' QA style in the train set.

Looking at the bottom number in each \emph{diagonal} cell from Figure \ref{fig:inter_annotator}, we see that our model performs quite differently on different annotators. Some annotators, such as w118 and w14, have a very high performance (90.0\% and 64.5\%, respectively), while some annotators, such as w24 and w313, have a relatively low performance (31.4\% and 24.6\%). This shows different annotator's questions have different level of biases.

We also discover that all annotators seem to transfer well to w118. We hypothesize w118 may have asked many questions that are similar to other annotator's questions which the model has already learned to answer during training time.

\paragraph{Dataset Re-split} The observation above incentivizes further investigation: what if we construct a re-split of the dataset where the validation set does not contain annotators in the train set? We conduct this experiment with the limited scope of the top-10 annotators used in Figure \ref{fig:inter_annotator} for clearer comparison. We create 11 re-splits of the dataset: 1 with annotator-overlapping train and validation set and 10 with annotator-non-overlapping train and validation set (use 9 annotators for train set and use 1 annotator for validation set). The results are shown in Table \ref{tab:non-overlap_resplit}. We find that 9 out of 10 for TVQA non-overlapping re-splits incur decrease of performance (less bias). Interestingly, the re-split where there is an increase in performance, w118, matches the columns in Figure \ref{fig:inter_annotator} whose cells' color is darker than average. This further verifies our explanation that w118 asks similar questions to other annotators. Nonetheless, this overall performance decrease trend after re-split suggests that for pretrained language models, annotator-non-overlapping re-split is a harder task than annotator-overlapping split and such re-split can help alleviate the QA bias. Based on this observation, we recommend future research work should create and use an annotator-non-overlapping split for train, validation and test sets whenever possible. The performance reported under such setting will contain fewer annotator biases and is thus a more accurate indicator of progress.

\subsection{Source of Bias: Question Type}

\begin{figure}[h!]
  \centering
  \includegraphics[width=.95\linewidth]{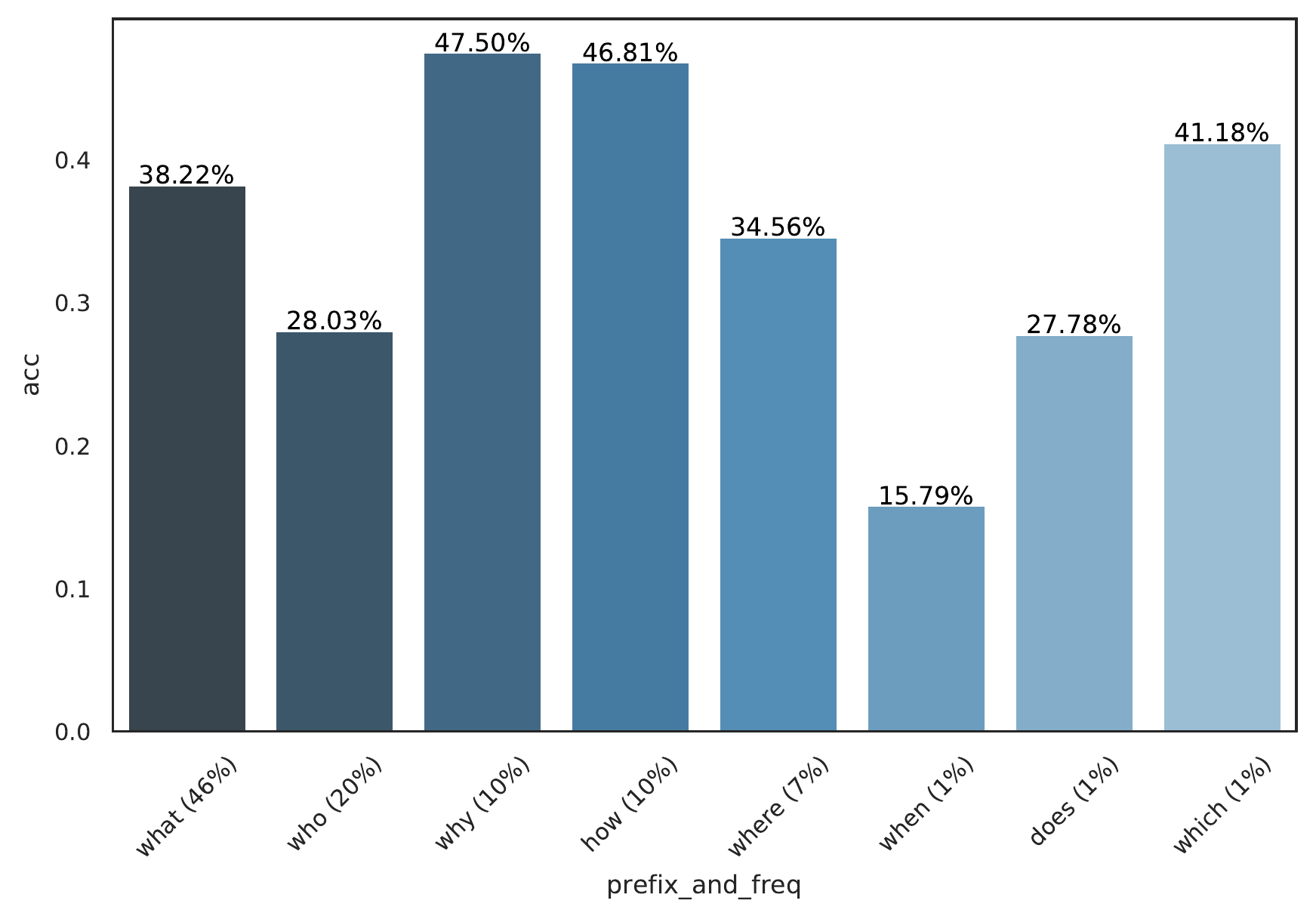}  
  \caption{MovieQA (A5) Accuracy by Question Type}
  \label{fig:movieqa_q_type}
\end{figure}

\begin{figure}[h!]
  \centering
  \includegraphics[width=.95\linewidth]{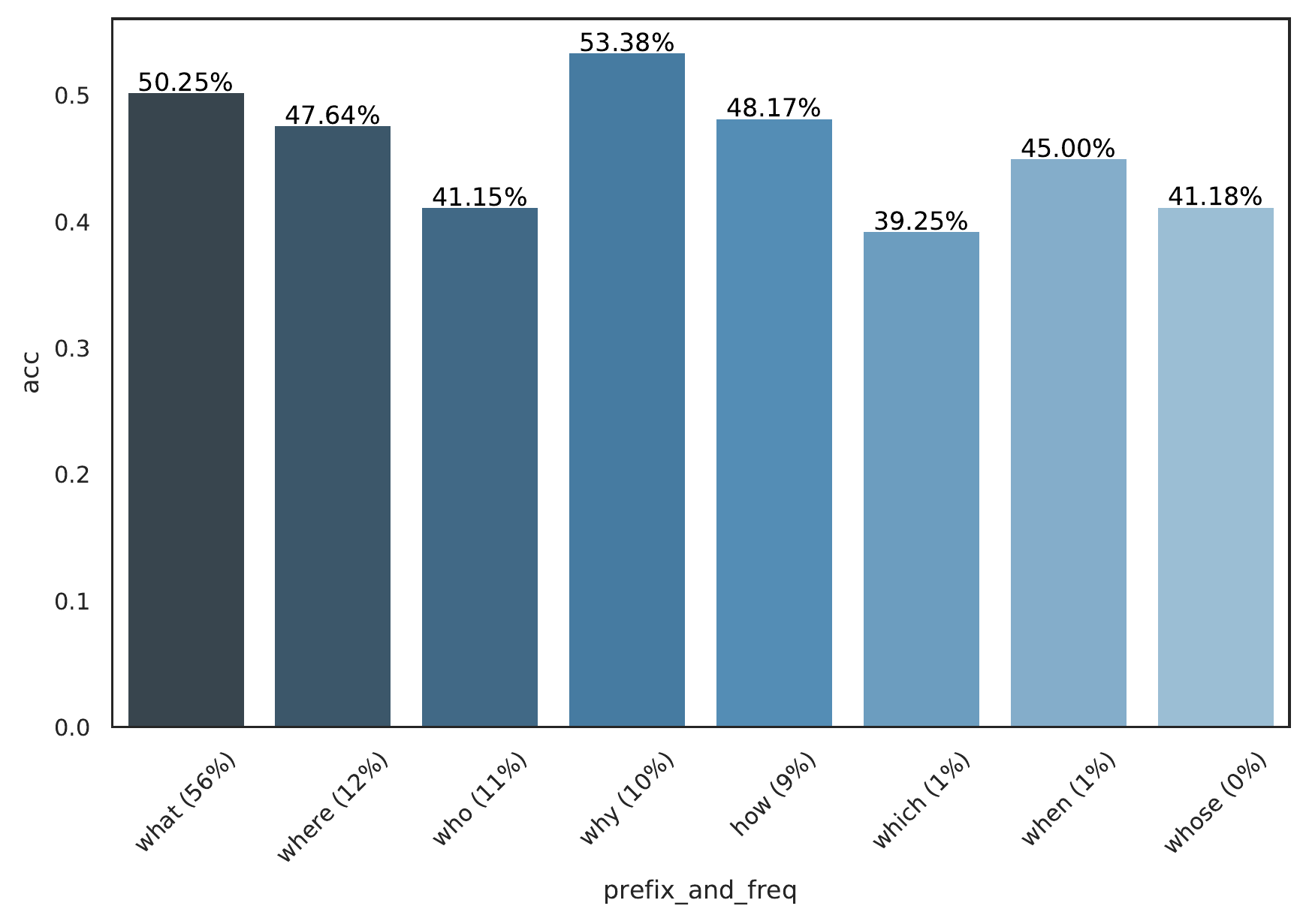}  
  \caption{TVQA (A5) Accuracy by Question Type}
  \label{fig:tvqa_q_type}
\end{figure}

We also hypothesize type of questions, such as reasoning question (such as why/how questions) vs. factual question (where/who questions), can be a source of bias. To verify, we ablate the model's accuracy based on the question's prefix. The results are shown in Figure \ref{fig:movieqa_q_type} and \ref{fig:tvqa_q_type}. These ablations are done on the A5 version of each dataset: Recall the random guess baseline in this case is 80\%: 20\%.

In Figure \ref{fig:movieqa_q_type}, we see MovieQA shows a clear distinction ($>10\%$) between ``why" ``how" questions vs. ``what", ``who", ``where" questions. The model fits significantly better to the former than the latter. 

In Figure \ref{fig:tvqa_q_type} for TVQA, the model can guess ``why" questions better than other question categories, while guessing ``who" remains difficult.

In general, we observe a trend that questions such as ``why" and ``how", which are reasoning and abstract questions and whose answers are more complex, incur more biases that language model can exploit; whereas ``what", ``who" and ``where" questions, which are factual and direct and whose answers are simple, are less bias-prone.


\section{Related Work}
Although more analysis \citep{goyal_making_2017, leibe_revisiting_2016} have been done on Visual Question Answering (VQA) \citep{Agrawal2015VQAVQ}, there are few works analysing biases in Video Question Answering datasets.  \citet{Jasani2019AreWA} suggest MovieQA contain biases by showing that about half of the questions can be answered correctly under the QA-only setting. However, their word embeddings are trained from plot synopses of movies in the dataset and thus they actually introduce context information into their model, making it no longer QA-only.
\citet{goyal2017making} propose that language provides a strong prior that can result in good superficial performance and therefore preventing the model from focusing on the visual content. They attempt to fight against these language biases by creating a balanced dataset to force the model focus on the visual information.
Similarly, \citet{cadene2019rubi} design a training strategy to reduce the amount of biases learned by VQA models named Rubi to counter the strong biases in the language modality.
\citet{manjunatha2019explicit} provide a method that can capture macroscopic rules that a VQA model ostensibly utilizes to answer questions.
However, those models fail to explain clearly where the bias in the dataset comes from, which is the main topic of our work.


\section{Conclusion}
In this work, we fine-tune pretrained language model baselines for two popular Video QA datasets and discover that our simple baselines exceed previously published QA-only baselines. These strong baselines reveal the existence of non-trivial biases in the datasets. Our ablation study demonstrates these biases can come from annotator splits and question types. Based on our analysis, we recommend researchers and dataset creators to use annotator-non-overlapping splits for train, validation and test sets; we also caution the community that when dealing with reasoning questions, we are likely to encounter more biases than in factual questions.

This paper is an post-hoc analysis for the datasets. However, the tools used in this paper could potentially also be extended to aid dataset creation. For example, a dataset creator could have a RoBERTa trained \emph{online} as annotators add more data. The annotators can use this language model's prediction to self-check if they are injecting any QA bias while coming up with the questions and answers. The dataset creator can also use a confusion matrix like Figure \ref{fig:inter_annotator} to monitor and identify low-quality annotators and decide the best strategy to reduce biases during the dataset creation process.

\section*{Acknowledgments}

We thank the anonymous reviewer for providing helpful feedbacks.


\bibliography{main}
\bibliographystyle{acl_natbib}

\appendix

\section{Appendices}
\label{sec:appendix}

\section{Supplemental Material}
\label{sec:supplemental}

\section{Model Settings and Hyperparameters}
We use the \texttt{roberta-large-mnli} checkpoint in the HuggingFace \texttt{transformers} GitHub Repo \footnote{\url{https://github.com/huggingface/transformers}} with the default hyperparameters. For the results reported in this paper, we use learning\_rate=$1\times 10^{-6}$ and batch\_size=3. Note that batch\_size=3 here means there are 3 \emph{questions} in one batch, along with all associated answers. All models are trained for 16 epochs and we take the last checkpoint to use for evaluation. This means that we treat validation set like test set: we do not do any hyperparameter search on the validation set.

\end{document}